\documentclass{llncs}
\usepackage{makeidx}  % allows for indexgeneration
\usepackage{url}      
\usepackage{subfigure}
\usepackage{amsfonts}
\usepackage{amsmath}
\usepackage{cancel}
\usepackage{graphicx}
\usepackage{color}
\usepackage{listings}
\usepackage{hyperref}
 
\usepackage{tikz}
\usepackage{ifthen}
\usepackage{xcolor}
\usepackage{multirow}
\usetikzlibrary{shadows.blur}

\newlength{\forkmeoffset}
\definecolor{forkmebg}{HTML}{CC0000}
\definecolor{forkmefg}{HTML}{EEEEEE}

\newcommand{\etal}{\textit{et al}.}

\newcommand{\forkme}[1][west]{
	\ifthenelse{\equal{#1}{east}}{%
		\tikzset{forkmerot/.style={rotate=-45}}
	}{%
		\tikzset{forkmerot/.style={rotate=45}}
	}
	\begin{tikzpicture}[remember picture, overlay]
	\node[forkmerot, shift={(0, -\forkmeoffset)}] at (current page.north #1) {
		\begin{tikzpicture}[remember picture, overlay]
		\node[fill=forkmebg, text centered, minimum width=50em, minimum height=3.0em, blur shadow, shadow yshift=0pt, shadow xshift=0pt, shadow blur radius=.4em, shadow opacity=50, text=forkmefg](fmogh) at (0pt, 0pt) {   \fontfamily{phv}\selectfont\bfseries \href{https://github.com/faustomilletari/TOMAAT}{Fork me on GitHub} };
		\draw[forkmefg!60, dashed, line width=.08em, dash pattern=on .5em off 1.5\pgflinewidth] (-25em,1.2em) rectangle (25em,-1.2em);
		\end{tikzpicture}
	};
	\end{tikzpicture}
}

\usepackage{graphicx}
\graphicspath{{pics/}{figs/}}
\begin{document}

\frontmatter          % for the preliminaries
\pagestyle{headings}  % switches on printing of running heads
\mainmatter              % start of your contributions

\title{TOMAAT: volumetric medical image analysis as a cloud service}
\titlerunning{TOMAAT: volumetric medical image analysis as a cloud service}  % 

\author{
 Fausto Milletari\inst{1},
 Johann Frei\inst{2},
 Seyed-Ahmad Ahmadi\inst{3}
}

\institute{NVIDIA \and Technische Universit\"at M\"unchen \and Ludwig-Maximilian Universit\"at M\"unchen}

\authorrunning{F. Milletari, J. Frei, A. Ahmadi}   

\maketitle              % typeset the title of the contribution

\forkme[east]

\begin{abstract}
Deep learning has been recently applied to a multitude of computer vision and medical image analysis problems. Although recent research efforts have improved the state of the art, most of the methods cannot be easily accessed, compared or used by either researchers or the general public.  Researchers often publish their code and trained models on the internet, but this does not always enable these approaches to be easily used or integrated in stand-alone applications and existing workflows.
In this paper we propose a framework which allows easy deployment and access of deep learning methods for segmentation through a cloud-based architecture. 
Our approach comprises three parts: a server, which wraps trained deep learning models and their pre- and post-processing data pipelines and makes them available on the cloud; a client which interfaces with the server to obtain predictions on user data; a service registry that informs clients about available prediction endpoints that are available in the cloud. These three parts constitute the open-source TOMAAT framework.
\end{abstract}

\section{Introduction}
\label{sec:intro}
In the last few years the scientific community has witnessed the raise of deep learning methods which have proved to be able to deliver state-of-the-art performances on a multitude of tasks from various domains. Drug discovery, genomics, natural language processing, computer vision and medical imaging are just a few examples of the fields where deep learning has been successfully employed. In particular, medical image analysis has been revolutionized by deep learning techniques for tasks such as segmentation, classification and detection with performances that sometimes surpassed those of humans. 

Differently than in classical computer vision, most of medical data is volumetric. Magnetic resonance imaging (MRI), computed tomography (CT), positron emission tomography (PET/SPECT), and even ultrasound (US) images are acquired in 3 or more dimensions in order to capture representative data about the human body, ultimately leading to better diagnoses and assessments.

Recent deep learning approaches have employed both 2D slice-by-slice \cite{wang2017automatic}, and 3D fully volumetric convolutional neural networks for segmentation \cite{milletari2016v,yu2017volumetric,chen2017voxresnet,cciccek20163d}. Various kinds of data manipulation have also been implemented by researchers as part of pre- and post- processing workflows needed to replicate the reported results.

Despite the popularity and the performances of recent methods, it is currently difficult to access state-of-the-art deep learning algorithms designed for medical tasks in order to run comparison, integrate them in existing workflows and use them for experimentation.
Some groups have provided implementations of their work \cite{milletari2016v,cciccek20163d,wang2017automatic}, others have provided the community with a framework containing previous approaches for comparison purposes and an open-source code-base enabling further development \cite{bonmati2017automatic}. Despite these efforts, most of the code and models available online do not follow a common standard and third-party implementations are not always correct. Moreover, due to the prohibitive hardware requirements in terms of memory and processing power of some methods, pre-trained models cannot be easily run by users, especially when deployment happens on outdated computing platforms. 

Bringing the efforts of medical image analysis researchers closer to the medical community for integration in existing clinical workflows and further investigation in clinical studies, is a priority that cannot be ignored any longer.

An important step in this direction is represented by DeepInfer \cite{mehrtash2017deepinfer} which proposes a strategy to make deep learning models running in docker containers available through a purpose-built module available as an extension of 3D Slicer \cite{fedorov20123d}. Differently than our framework, algorithms made available by DeepInfer are executed on the local machine and not on the cloud: DeepInfer requires a local docker installation on the client machine as well as high performance hardware to be available. 

In this paper we propose TOMAAT, an open-source framework and architecture that allows to (i) enable researchers to make their algorithms accessible by packaging their models and data processing pipelines into a prediction service with an unique interface; (ii) enable users to access prediction services through a client that is simple and straightforward to adopt and operate. In this way we facilitate use, comparison and further development of 3D segmentation approaches; (iii) enable researchers to add their algorithms to a public algorithm list that can be used by users to select the algorithm they want to leverage to obtain predictions. 

All algorithms are deployed on the network and can be accessed by any client complying with our interface. We provide an example of a client as a module of 3D Slicer. Any computer having the capabilities to run 3D Slicer can use our predictions endpoints without any requirement for additional software or computational resources to be installed.  

We hope in this way to speed up and improve research by:
(i) making different approaches directly comparable such that their advantages and pitfalls can be adequately understood and highlighted; (ii) disseminating scientific work in a more straightforward manner by allowing users to access a list of current segmentation methods for each modality/anatomy; (iii) allowing users to access state-of-the-art approaches and use them to obtain results on their own data. We hope in this way to bridge the gap between the medical and computer science communities; (iv) allowing portability of models. Once wrapped using the paradigm proposed by TOMAAT, they can be packaged in containers and run anywhere using the same interface; (v) making models accessible in exactly the same way regardless of whether they are deployed on a local machine, a private network or the cloud;

The homepage of the TOMAAT project can be accessed at the address \newline \url{http://tomaat.cloud} while the relative code repositories can be found on GitHub.

\section{Method}
\label{sec:method}
\begin{figure} 	
\centering 	
\includegraphics[scale=0.20]{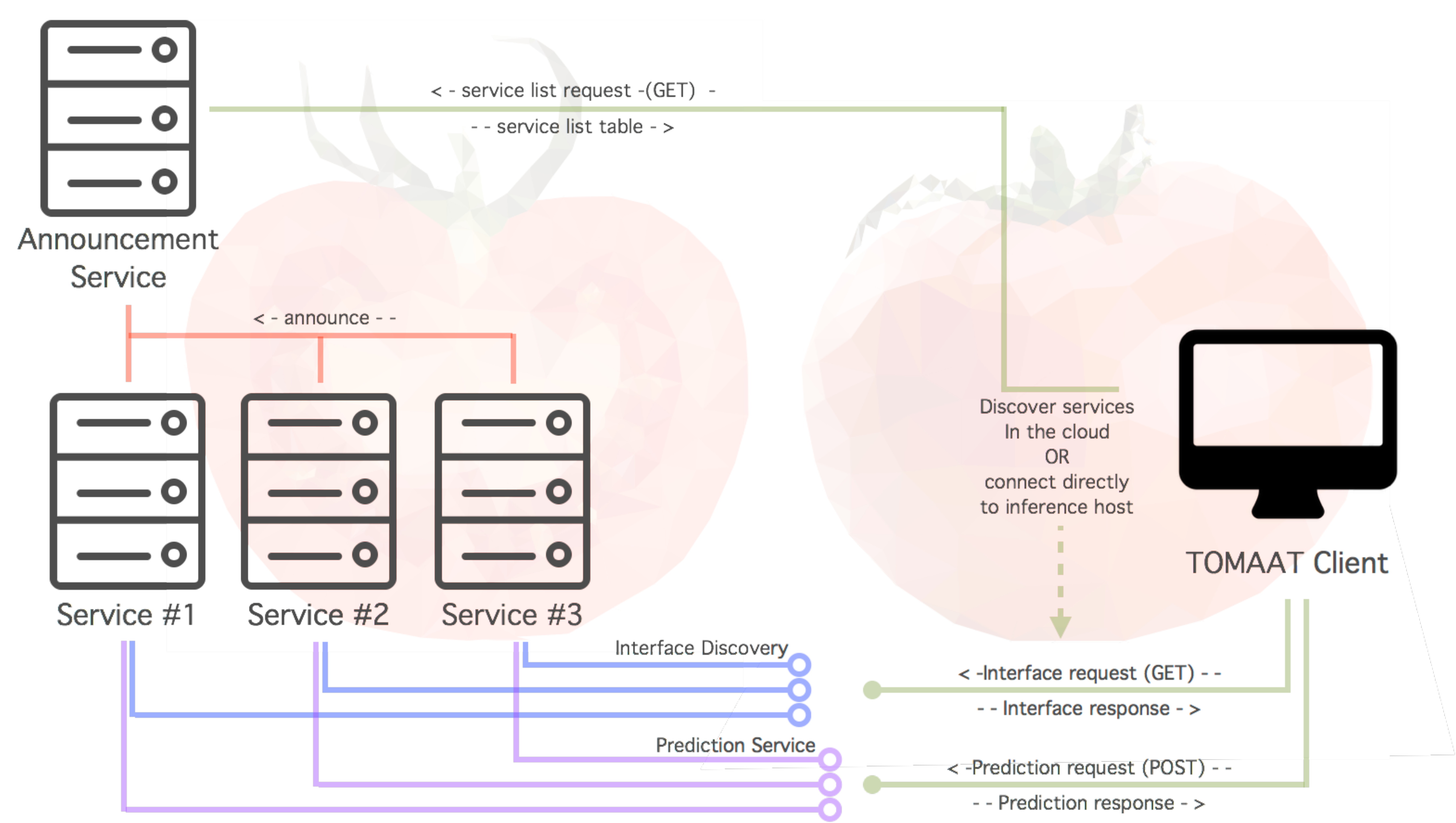} 	
\caption{Schematic representation of the architecture of the TOMAAT framework.} \label{fig:architecture} 
\end{figure}

In this section of the paper we describe the approach followed for the implementation of TOMAAT. A summary of this approach can be seen in Figure \ref{fig:architecture}. The system can be divided in three parts: server side, client side and announcement service.

We define "server" or "prediction endpoint" a machine equipped with adequate hardware which runs a deep learning model and its pre- and post-processing pipelines wrapped by TOMAAT. A server is a prediction endpoint in the network, meaning that it can be accessed by clients to obtain predictions through a network connection (local or remote) using the interface provided by TOMAAT and the HTTP protocol. Users can choose whether they want to make their servers public or not by announcing them to the "announcement service".

We define "client" a machine equipped with TOMAAT client which is currently implemented in a 3D Slicer \cite{fedorov20123d} extension that can be downloaded using the Slicer extension manager. 
The client allows straightforward usage of remote and local prediction endpoints. Results can be stored to disk by saving them using the 3D Slicer user interface. Alternatively, it is possible to interact with servers by sending them appropriate HTTP requests.

We define "announcement service" a system that maintains an updated list of all the public servers that have been registered as valid prediction endpoints capable of running segmentation of specific anatomies in specific modalities. Communications between clients, servers and announcement service is realized through HTTP as well. 

\subsection{Server}
\label{sec:Server}
Once a deep learning model has been trained, its whole functionality at test time can be subdivided in three parts: pre-processing, inference and post-processing. 

During pre-processing the data needed for computation is read from the disk and transformed such that it matches the format and characteristics expected by the neural network.
During inference the neural network model is run. Once the pre-processed data is presented as input to the network, the corresponding outputs can be obtained by executing the forward step. 
Finally, during post-processing, a series of transformations are applied to the outputs of the neural network in order to refine the results and convert them to the expected output format. 

The model functionality is wrapped by our server which implements a communication protocol over HTTP 1.1 in order to allow clients to request and obtain predictions.
TOMAAT allows to define pre- and post- processing functions as well as custom prediction functions leveraging any deep learning framework. The idea is that the user should be responsible for defining the inference function, as well as the transformations that happen on the data. We provide, together with our package, an example of such an implementation that can also be used to serve generic Tensorflow models. 

Depending on the task and on the approach used, the number and type of arguments required to run the forward step may vary. In order to support any typology of model and task, TOMAAT server is designed to allow researchers to define custom interfaces for their services. That is, the Servers HTTP-based interface can be completely customized. Interfaces in are defined in our framework in terms of standard elements which can be used by the client as instructions on how to build appropriate messages that can be used and understood during prediction. 
For example, a method that requires multiple input volumes as an input, communicates to the client its needs before the actual prediction request takes place 
A list of elements that need to be included in each request can be obtained through a GET request to a specific URL served by the prediction endpoint. 

Predictions endpoints are powered by the python package "Klein". "Klein" is "is a micro-framework for developing production-ready web services with Python" \cite{Klein}. It has the ability of running request handlers in threads, which enables multiple clients to be served at the same time. In the implementation we provide on GitHub, however, threads compete for the usage of the GPU through a locking mechanism.

Another characteristic of our framework is the ability of handling algorithms having different aims, ranging from classification to detection and segmentation. To enable this we must be able to return different data types as prediction result. We handle this situation, by creating response messages that contain a brief description of the returned data withing themselves. An arbitrary number of data-fields belonging to a few standard types can be returned as a result of the prediction.

\subsection{Announcement service}
As previously discussed, prediction endpoints can be published to an announcement service which maintains a list of available servers in the network. A server needs to announce itself through a POST request containing essential information about the prediction endpoint itself such as its URL, description, modality, task, anatomy, and other information as per documentation. Crucially, in order to successfully announce a new prediction endpoint, a secret API-key, unique to each serve, must be supplied with the POST request. In this way we limit the capability of announcing services only to trusted developers.

The list of currently available services can be retrieved by a GET request to the announcement service. Information about currently available servers will be returned to the client together with an unique identifier for each server, such that clients that are built to communicate with a specific prediction endpoint (e.g. are part of a workflow and do not support user interaction) can select the remote server using that identifier.

\subsection{Client}
Clients can communicate with servers through POST requests. They can be implemented in any language and integrated in any platform as long as they implement the interface of TOMAAT. Clients aiming to be general purpose and aiming to interact with any prediction endpoint, must be capable of handling the element-based interface specified by each server. This can be done by creating a modular interface that support the presence of multiple elements. Similarly, it is necessary to implement the capability of handling modular response messages containing different data format and information. 

\begin{figure} 	
\centering 	
\includegraphics[scale=0.28]{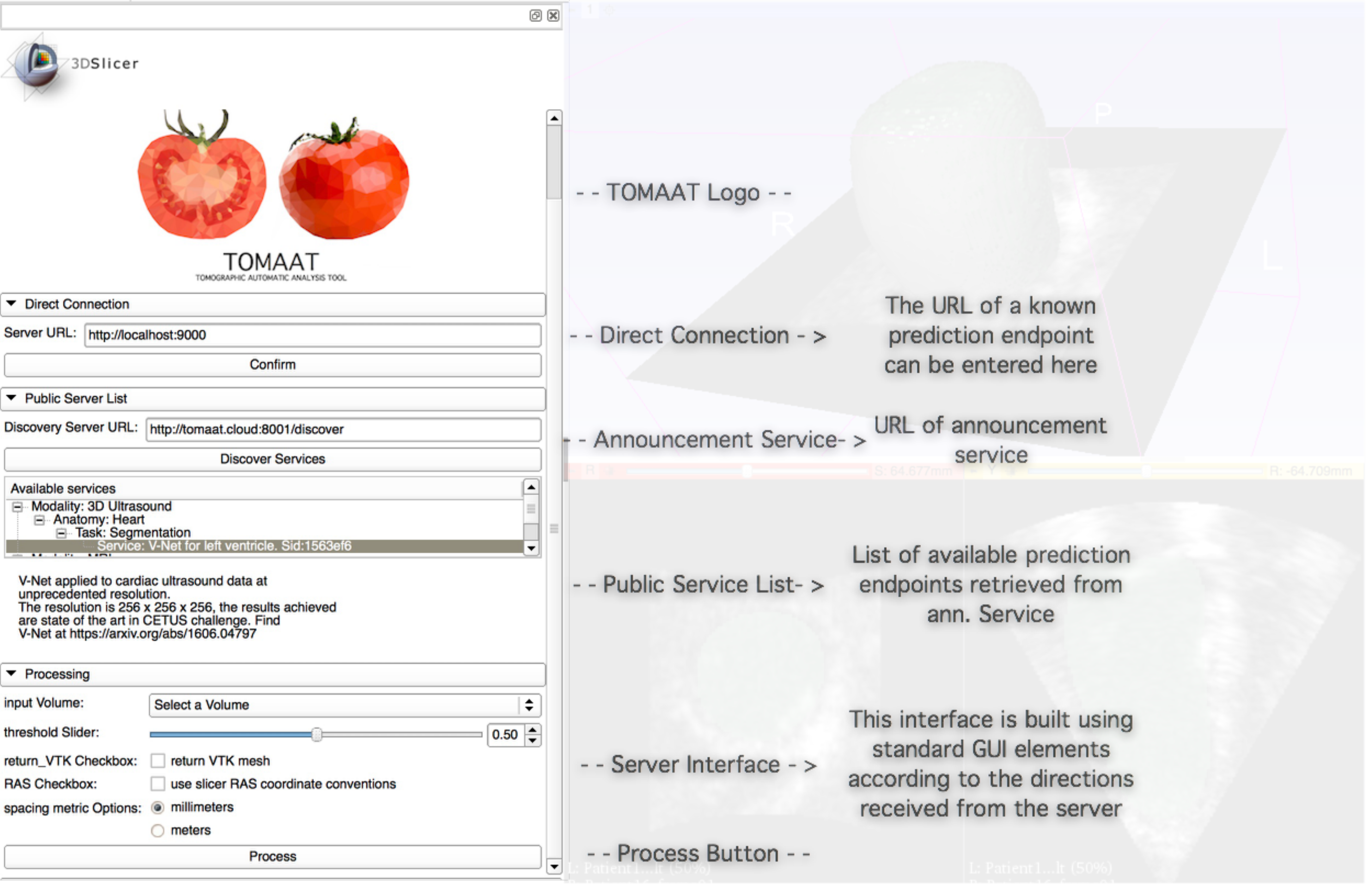} 	
\caption{3D slicer module allowing interaction with TOMAAT announcement service and prediction endpoints.} \label{fig:module} 
\end{figure}

We have developed a general-purpose client module with these characteristics (Figure \ref{fig:module}) for 3D Slicer \cite{fedorov20123d}. Through this module we allow users to interact with our framework. We allow clients to (i) realize direct connections to servers whose URL (local or remote) is already known, to (ii) obtain the list of currently available public servers from any announcement service (default \url{http://tomaat.cloud:8001/discover}), (iii) obtain results from prediction endpoints.

After selecting a server via either the "Direct connection" or "Public server list" pane, the module requests information about the interface to the server. At this point, the content of the "Processing" pane of the user interface is built according to these instructions. A number of GUI widgets is created to allow user interaction. When the "Processing" button is clicked a prediction request is made. When a response is received, each piece of returned data is handled according to its format, as explained in Section \ref{sec:Server}.

\section{Experimental evaluation}
\label{sec:results}

In order to demonstrate the capabilities of our framework, we tested three different prediction endpoints, dealing with three segmentation tasks. The first server deals with segmentation of 3D ultrasound data of the heart, and aims to segment the left-ventricle using a 3D FCNN inspired by V-Net \cite{milletari2016v}; the second service segments prostates in MRI volumes using another network inspired by V-Net \cite{milletari2016v}; the third consists of the work of Wang \etal, presented in \cite{wang2017automatic}, aiming at brain tumor segmentation. 

The left ventricle segmentation use case takes advantage of a neural network operating at a very high volume resolution of 256 cubic voxels; the prostate segmentation application uses much less GPU memory and compute power by processing inputs that are 256x256x64 voxels; finally, the brain tumor segmentation algorithm leverages multiple 2D networks operating on the volumes slice by slice in three directions as explained in \cite{wang2017automatic}. The computation and post-processing steps of this method are heavy. This method also requires the user to transfer to the server 4 MRI volumes instead of one. This is due to the fact that Flair, T1, T1c and T2 volumes need to be processed to obtain the final segmentation.

In this experimental section we compare the performances of TOMAAT in terms of response latency time in three scenarios: when the server is available through a LAN connection, when a DSL-like connection is used and when the prediction is requested through a 4G mobile connection. For the first two methods, we also report the net GPU compute time (which does not include pre- and post- processing), as baseline. The results are shown in Table \ref{tab:results}.

\begin{table}
\caption{TOMAAT latency for different scenarios.}\label{tab:results}
\begin{tabular}{|c|c|c|c|}
\hline 
Method & LAN latency & DSL latency & 4G latency \tabularnewline
\hline
\hline
3D V-Net Heart& 3.93s (GPU: 1.25s) & 6.21s (GPU tme: 1.24s) & 8.70s (GPU: 1.25s) \tabularnewline
\hline 
3D V-Net Prostate & 3.40s (GPU: 0.38s) & 7.42s (GPU: 0.38s) & 13.53s (GPU: 0.389s) \tabularnewline
\hline 
Brain tumor & 20.12s & 41.16s & 82.49s \tabularnewline
\hline 
\end{tabular}

\end{table}

\section{Conclusion}
In this paper we have presented TOMAAT, an open-source framework to deploy and access deep learning models for 3D medical image analysis in an easy and straightforward manner. Our aim is to speed up and improve research in this field by giving the possibility to users to experiment, compare and use state-of-the-art deep learning models for tomographic segmentation. Our framework allows to deploy models as cloud services accessible through a simple HTTP interface. Models can be made public by communicating the URL of the prediction endpoint to an announcement service, and can in this way be accessed by users.

\bibliographystyle{splncs03}
\bibliography{bibliography}

\end{document}